\documentclass{article}
\usepackage{spconf,amsmath,graphicx}
\usepackage{color}
\usepackage{algorithm}
\usepackage[noend]{algpseudocode}

\algnewcommand{\LineComment}[1]{\State  \(\triangleright\) #1}
\usepackage{amsmath}
\usepackage{amssymb}
\usepackage{multirow}

\title{Instance-based model adaptation for direct speech translation}

\name{Mattia A. Di Gangi$^{1,2}$, 
Viet-Nhat Nguyen$^1$\sthanks{Work done during a Summer internship at FBK.}, Matteo Negri$^2$, Marco Turchi$^2$}
\address{$^1$University of Trento~~~~~~~~~~~~~$^2$Fondazione Bruno Kessler\\
	Trento, Italy}
\begin{document}
%
\maketitle
\begin{abstract}
Despite recent technology advancements, the effectiveness of neural approaches to end-to-end speech-to-text translation is still limited by the paucity of publicly available training corpora. We tackle this limitation with a method to improve data exploitation and boost the system's performance at inference time. Our approach allows us to customize ``on the fly'' an existing model to each incoming translation request. At its core, it exploits an instance selection procedure to retrieve, from a given pool of data, a small set of samples similar to the input query in terms of latent properties of its audio signal. The retrieved samples are then used  for an instance-specific fine-tuning of the model. We evaluate our approach in three different scenarios. In all data conditions (different languages, in/out-of-domain 
adaptation),
our
instance-based adaptation yields coherent  
performance gains over  static models.

\end{abstract}


\section{Introduction}
\label{sec:intro}

The technology advancements in end-to-end speech-to-text translation (ST) recently allowed to reduce the performance gap with classic cascade solutions combining separate automatic speech recognition (ASR) and machine translation (MT) components. 
However, despite its advantages in terms of architectural simplicity and reduced error propagation, direct ST still suffers from drawbacks related to its 
limited data effectiveness \cite{sperber2019attentionSHORT}. 
A general problem is that neural approaches are \textit{per se} data-hungry and the publicly available ST corpora are still orders of magnitude smaller than those released for ASR and MT \cite{mustc19}. 
The data demand issue is exacerbated by the fact that, being a higher-level task than ASR and MT,  direct ST  requires higher abstraction capabilities to capture relevant features of the input (audio signals) and learn  the mapping into proper output  representations (texts in the target language). 
Learning this mapping  end-to-end is usually more complex and data demanding than exploiting the intermediate representations of separate, individually trained components.

Previous solutions to cope with data scarcity focused on two orthogonal aspects: improving the learning process and increasing the training material.
On the learning side, \cite{kano2017structured,weiss2017sequence,berard2018end,anastasopoulos2018tied,bansal2018pre,di2019enhancing} exploited transfer learning 
from ASR and MT showing, for instance, that pre-training the ST encoder on ASR data can yield significant 
improvements.
On the data side, the most promising approach is data augmentation, which has been experimented via knowledge distillation from a 
neural MT (NMT) model \cite{liu2019end}, 
synthesizing monolingual MT data in the source language
\cite{jia2018leveraging},
multilingual training \cite{digangiASRU2019}, 
or 
translating monolingual ASR data into the target language \cite{jia2018leveraging,digangi2019data,liu2018ustc}. 
Nevertheless, despite 
some claims of big industrial players operating in rich data conditions
\cite{jia2018leveraging},
top results at recent 
shared tasks
\cite{liu2018ustc} show that effectively exploiting the scarce training data available still remains a crucial issue to 
reduce the performance gap with cascade ST solutions. 

Along this direction, we propose a general framework 
for maximizing data exploitation and customizing  an existing ST model to each incoming translation request at inference time.
In a nutshell, given a generic model  $M_{g}$ and 
an ST data pool $D$, each translation request $r$
is handled by a two-step process. First, a set of (\textit{audio}, \textit{translation}) pairs is retrieved  from $D$ based on the similarity between their  audio element and 
$r$.
Then, the retrieved pairs are used to adapt $M_{g}$ via fine-tuning.
The 
underlying 
intuition is that the similarity of the new samples with the input audio can be used at run-time to 
overfit $M_{g}$ 
to samples similar to $r$, and influence its behaviour towards a better translation.

We explore this idea in 
different scenarios 
considering different language directions and experimenting 
in intra-, multi- and cross-domain adaptation.
Our results show that, compared to 
static ST models, 
instance-based on-the-fly adaptation 
 yields variable but coherent improvements, with larger gains in cross-domain scenarios where the 
mismatch between the training and test domains makes 
ST 
more 
challenging.

\section{Direct Speech Translation}
\label{sec:task}

In direct speech translation, a single neural model is trained end-to-end on the speech-to-text translation task. Given an input audio segment $\mathbf{X}$ representing a speech in a source language $e$, and an output text $\mathbf{Y}$ representing the translation of $\mathbf{X}$ in a target language $f$, a direct ST model is trained by optimizing the log-likelihood 
function in Equation 1, 
where $B$ is the size of a batch, $l_b$ is the length of the target sequence at position $b$, and $\mathbf{\theta}$ is the vector of model's parameters.

\begin{equation}
    L = -\sum_{b=0}^B \sum_{i=0}^{l_b} y_{ib} \log(p(\tilde{y}_{ib} \vert \mathbf{X}, y_{< i,b}; \mathbf{\theta}))
\end{equation}

Models for this task have a sequence-to-sequence architecture \cite{sutskever2014sequence} with at least one encoder that processes the audio input, and one decoder that generates the output, one token at a time, in an autoregressive manner. 
In this work, we  use  S-Transformer \cite{digangi2019adapting}, an adaptation of  Transformer~\cite{vaswani2017attention} to the ST task. 
In addition to the original Transformer,  S-Transformer elaborates the input spectrograms with 
\textit{ad-hoc} layers. The input is first processed by two stacked 2D CNNs with stride (2, 2), which also reduce the input sequence length by a factor of $4$. Then, the output of the second CNN is fed to a stack of two 2D Self-Attention~\cite{dong2018speech}. The goal of the 2D Self-Attention is to model the bi-dimensional dependencies along the spectrogram's time and frequency dimensions. 
2D Self-attention layers process the input with 2D CNNs and compute attention along both matrix directions. 
All CNNs are followed by batch normalization~\cite{ioffe2015batch} and ReLU nonlinearity. Moreover,  to  focus the  encoder  on  short-range dependencies, a distance penalty mechanism is added in  every  self-attention layer of the encoder. Given a position $i$ in the query vector, and a position $j$ in the key vector, with $i \neq j$ we compute $\textit{pen}=\log(\vert i - j \vert)$ and subtract \textit{pen} from the attention scores before softmax normalization.

\section{Instance-based Model Adaptation}
\label{sec:method}

Algorithm~\ref{algo:uniad} illustrates our 
instance-based model adaptation procedure. Its goal is to improve the performance of a pre-trained ST model $M_{g}$ by fine-tuning it at inference time on (\textit{audio}, \textit{translation}) pairs in which the audio is similar to the input translation request $r$. These pairs are retrieved from a data pool $D$, which can either be 
the same training set used for $M_{g}$ or a new dataset.
In the former case, instance-based
adaptation aims to maximize the exploitation of the training data.
In the latter case, the goal is to exploit newly available data to also cover new domains.
Our experiments ($\S\ref{sec:experiments}$) will address both the scenarios.

\begin{algorithm}[h!]
\small
\caption{Instance-based Model Adaptation (IMA)}\label{algo:uniad}

\begin{algorithmic}[1]
\LineComment $M_{g}$: generic ST model 
\LineComment $M_{r}$: adapted ST model 
\LineComment $D$: ST data pool
\LineComment $r$: translation request
\LineComment $\tau$: similarity threshold
\LineComment $D_{r}$: $\{(a_1, t_1), ..., (a_n, t_n)\}$ retrieved (audio, translation) pairs
\LineComment $t^{*}$: translated segment
\Procedure{IMA($M_{g}$, $D$,  $r$, $\tau$)}{}
    \LineComment Local copy of the generic model
    \State $M'_{g}$:=$M_{g}$
    \LineComment Instance selection
    \State $D_{r}$:={\bf Retrieve}($r$, $D$, $\tau$)
    \If{$D_{r}$  $\not=$  $\varnothing$}
        \LineComment Model optimization
        \State $M_{r}$:={\bf Adapt}($M'_{g}$, $D_{r}$)
    \Else 
        \State $M_{r}$:=$M'_{g}$
    \EndIf
    \LineComment Translate the segment with the adapted ST model
    \State $t^{*}$:={\bf Translate}($M_{r}$, ${r}$)
    
\EndProcedure
\end{algorithmic}
\end{algorithm}

\noindent
\textbf{Data pool.}
$D$ consists of \textit{(audio, translation)} pairs, in which 
the audio element is also used as a 
retrieval key for
the pair. 
For our experiments, the audio segments are stored 
either  \textit{i)} as a spectrogram $S$ with $N$ time frames and $k$ features (Raw Features in Table \ref{tab:intra}), or \textit{ii)} as a function $E(S)$ obtained by processing $S$ with the model's encoder (Encoder Features). The generated segments are stored in order to be retrieved during translation.

\noindent
\textbf{Similarity.}
The similarity between the query audio segment $r$ and the 
audio segments in $D$
is computed as the cosine similarity between the pairs of vectors 
$(z_r, z_1),  \dots, (z_r, z_n) \in R^{k}$, 
where $k$ is the number of features of the chosen segment representations. Each $z_i$ is obtained by summing all the time frames of its sequence along the time axis.
The advantage of this similarity is its applicability in a direct ST scenario, where no intermediate transcription step is involved.

\noindent
\textbf{Retrieval.} 
The retrieval procedure receives as argument the translation request $r$, 
the data pool
$D$ and a similarity threshold $\tau$. It returns the set of (\textit{audio}, \textit{translation}) pairs ($D_{r}$) for which the similarity of the audio element with $r$ is above $\tau$.

\noindent
\textbf{Adaptation.} 
If $D_{r}$ is not empty, the generic model $M_g$ is  fine-tuned for $e$ epochs on the top $n$ samples to obtain the adapted model $M_r$ used to translate $r$. In our experiments, $e$ and $n$ are fixed hyperparameters. After translating $r$, the adapted model is discarded 
so that, for the next input query, the process restarts from the initial generic  model $M_g$.

\section{Experiments}
\label{sec:experiments}

\subsection{Datasets}
\label{ssec:dataset}

We use two datasets. One is MuST-C \cite{mustc19}, a multilingual ST corpus containing English speech (TED Talks) translated into 8 European languages. Data size ranges from 385 hours for English$\rightarrow$Portuguese to 504 hours for English$\rightarrow$Spanish. 
The other corpus is How2 \cite{sanabria2018how2}, a multimedia corpus for English$\rightarrow$Portuguese also including  ST data (300 hours). 
In both corpora, the speech segments are in the form of log MEL filterbanks with time width 25ms and step of 10ms. 

A comparison between the target side of the En-Pt section of MuST-C and How2  shows that they have a different level of 
text repetitiveness (How2 has a repetition rate \cite{cettolo2014repetition} that is 40\% higher) and vocabulary overlap (27\% of the MuST-C terms appear in How2, while 48\% of the How2 terms are also in MuST-C).
Other differences in terms 
background noise 
and number of non-native speakers (both higher in MuST-C) suggest  
that the How2 data are in general easier to handle 
for ST training/adaptation.
Depending on the selected test set, we hence expect variable 
gains over the static ST models.

\subsection{Settings}
We trained S-Transformer on all the datasets with the following hyper-parameters: 2D CNNs have kernels of size $3\times 3$ and stride (2, 2), 2D self-attentions have internal 2D CNNs with $4$ output channels (and thus $4$ heads in multi-head attention), and $64$ output channels in the last layer. Transformer layers have size $512$ with $8$ heads in multi-head attention and $1024$ units in the hidden feed-forward sub-layers. Dropout is set to 0.1 after each layer.
For training, we used the Adam optimizer \cite{kingma2014adam} with noam decay \cite{vaswani2017attention} using initial learning rate 0.0003, $4000$ warm-up steps and maximum learning of $0.001$. The loss we used is cross-entropy with label smoothing \cite{szegedy2016rethinking} set to $0.1$. The batch size is of 4 segments, but we trained on 4 GPUs NVIDIA K80 and accumulated gradients for $16$ batches.
Target texts are split at character level. The results are 
computed using the BLEU score \cite{papineni2002bleu} at word level.

\subsection{Experiments}
\label{ssec:experiments}

\begin{table}[t]
\centering
\small
\begin{tabular}{l|c|c|c}
  \multicolumn{4}{c}{\textbf{\textit{Intra-Domain}}}\\\hline
& \textbf{Baseline} & \textbf{ Raw Features} & \textbf{Encoder Features} \\\hline
De & 17.0 & 16.9 & \textbf{17.3} \\
Es & 21.5 & 21.5 & \textbf{22.0} \\
Fr & 27.0 & 27.1 & \textbf{27.4}\\
It & 17.5 & 17.8 & \textbf{18.0}\\
Nl & 21.8 & 21.9 & \textbf{22.0} \\
Pt & 21.5 & 21.4 & \textbf{21.7} \\
Ro & 16.4 & 16.4 & \textbf{16.8}\\
Ru  & 12.2 & 12.3 & \textbf{12.4}\\\hline
How2 & 39.4 & 39.9 & \textbf{40.1} \\
\end{tabular}
\caption{BLEU results on MuST-C and How2 in the intra-domain scenario. The retrieval is based on either MEL filterbanks or the encoder's output representations. }

\label{tab:intra}
\end{table}

We evaluate our instance-based model adaptation approach in three scenarios. In the first scenario (``\textbf{intra-domain}''), the data pool used for retrieval ($D$) is the same 
corpus used to train the initial ST model $M_{g}$.
These experiments aim to evaluate whether
instance adaptation helps 
to make better use of the training data.
In the second scenario (``\textbf{multi-domain}''), 
$M_{g}$ is trained on data from two domains ($D1$+$D2$) and the goal is 
to maximize 
performance on both. In this case, 
the adaptation is performed using as a data pool either the 
domain-specific material from the same domain of the query $r$, or the whole data from the two domains.
In the last scenario (``\textbf{cross-domain}''),
$M_{g}$ is trained  on data from one domain only, and it has to be adapted to a new domain. We consider two variants of this scenario. 
In the first variant, 
an in-domain data pool from the same domain of the test set is available for retrieval.
In the second variant, 
the data pool contains only the original, out-of-domain training data. 
The latter variant, in which 
$M_{g}$ has to be adapted to unseen test data by only exploiting out-of-domain material, represents the hardest condition from an on-field 
deployment standpoint.

\begin{table}[t]
\centering
\small
\begin{tabular}{l|c|c|c}
 &\textbf{1 Epoch} & \textbf{3 Epochs} & \textbf{5 Epochs} \\\hline
De  & 15.8 & 13.1 & \textit{10.0} \\
Es & 21.0 & 19.8 & \textit{19.0} \\
Fr &  26.3 & 22.5 & \textit{18.8} \\
It &  17.0 & 15.0 & \textit{13.2} \\
Nl &  21.5 & 20.1 & \textit{18.0} \\
Pt & 21.2 & 19.6 & \textit{17.7} \\
Ro &  15.8 & 13.7 & \textit{11.4}\\
Ru &  11.9 & 9.7 & \textit{7.3} \\
\end{tabular}
\caption{BLEU results on MuST-C running the adaptation for 1, 3 and 5 epochs on the 
least similar pair retrieved from $D$.}

\label{tab:firstlast}
\end{table}

For each setting, we perform hyperparameter search in the validation set, then
the best selection is applied on the test set. 
We perform 
instance-based 
adaptation with the Adam optimizer \cite{kingma2014adam} and choose the best set of hyperparameters among learning rates=$\{1, 2, 3\}\times 10^{-\{3,4,5\}}$, number of retrieved samples = $\{1, 5, 10\}$, and number of tuning epochs = $\{1, 3, 5\}$. Additionally, we filter out the retrieved samples whose cosine similarity score is below a threshold 
$\tau$.
After an initial exploration, we found out that a threshold 
$\tau=0.5$
allows the systems to keep the best performance while reducing the tuning time. As an additional note, we found that the SGD optimizer does not work as well as the Adam optimizer, particularly for the 
multi/cross-domain adaptation experiments.

\section{Results}
\label{sec:results}

\textbf{Intra-domain.}
The results of the intra-domain experiments are 
shown
in Table \ref{tab:intra}. In general, the performance  
on MuST-C is lower than 
on How2.
As pointed out in $\S\ref{ssec:dataset}$, despite the smaller 
size of the training corpus, 
the higher repetitiveness of How2 creates
a favourable evaluation condition. 
Instance-based adaptation, however, 
 provides 
small but coherent  improvements on all the language pairs and on both corpora (from 0.2 to 0.5 for MuST-C and 0.7 for How2). 
Since the Encoder Features are slightly better than the Raw Features, they will be used in the rest of the experiments.
To better understand the
effectiveness of our approach, Table \ref{tab:firstlast} 
shows the impact of adapting on the 
least similar pair retrieved from the pool, for different numbers of epochs and for each  language direction of MuST-C. These results 
are always worse
than the 
baseline and, by  increasing the number of epochs, they 
deteriorate  up to  $-7.5$ BLEU points on Fr with 5 epochs.
This suggests that  instance-based adaptation 
is sensitive to the quality (i.e. the similarity) of the 
retrieved material and that our approach is able to identify pairs that are useful to the model, 
resulting in variable performance 
gains in all the experiments.

\noindent\textbf{Multi-domain.}
To evaluate 
instance-based adaptation 
in the 
multi-domain scenario,
we trained our initial model ($M_{g}$) on the concatenation of the En-Pt data from MuST-C and the How2 data. The results presented in 
lines 1-3
of Table \ref{tab:cross-domain} 
indicate that using more data is beneficial for both the generic (+1.2 on the MuST-C baseline 
reported in Table \ref{tab:intra}
and +1.6 on 
How2) and the instance-based adaptation (+2.1 for MuST-C and +2.4 for How2). 
This can be explained by the fact that, when a model has been trained on larger and more diverse data,  
it is 
stronger due to its higher generalization capability.
In this case, instance-based adaptation can account for the domain shift without 
performance loss in the initial domains.

\noindent
\textbf{Cross-domain.}
As mentioned in $\S$\ref{ssec:experiments}, we also run our domain-adaptation experiments by training the ST model in one domain and testing it on the other. The similar pairs can be retrieved either from
the same 
domain of the test set 
or from
the training data only.
In general, when 
training and test 
data
come from different domains (Table \ref{tab:cross-domain}, line 4), 
the 
non-adapted models show a significant drop in performance (-11.6 BLEU points for the MuST-C test set and -25.3 for How2). 
Retrieving from the same domain (line 5) helps with gains over the static model of 1.2 BLEU points for the MuST-C test set and +7.5 for How2. These results are promising but still far from the baseline values reported in Table \ref{tab:intra}. 
However, it
is important to remark that 
our baselines have access to the in-domain data in advance, so they work in a more favorable condition. For the sake of comparison, we fine-tuned the 
baseline models on the incoming pool of in-domain data, but this results in models with performance comparable to the baselines for the new domain without pre-training.
Retrieving similar pairs from a different domain (line 6) is extremely difficult, in particular considering the differences between the two datasets (see $\S$ \ref{ssec:dataset}).
Also in this case, however,
instance selection is able to leverage the training data to produce translations that are slightly better than those obtained from the static system ($+0.6$ on MuST-C and $+0.3$ on How2).

\begin{table}[t]
\centering
\small
\begin{tabular}{l|c|c|c|c|c}
& \textbf{Train}  & \textbf{Test}  &  \textbf{Pool}  & \footnotesize{\textbf{D1: How2 }}&  \footnotesize{\textbf{D1: MuST-C}} \\
 &  &   &    & \footnotesize{\textbf{D2: MuST-C}}&  \footnotesize{\textbf{D2: How2}} \\\hline
  \multicolumn{6}{c}{\textbf{\textit{Multi-Domain}}}\\\hline
1 & D1 + D2 & D1  & -  &  22.7 & 41.0  \\
2 &  D1 + D2 & D1  & D1  &  \textbf{23.6} & \textbf{41.8}  \\
3 &  D1 + D2 & D1  & D1+D2  &  \textbf{23.5} & \textbf{41.8}  \\\hline \hline
  \multicolumn{6}{c}{\textbf{\textit{Cross-Domain}}}\\\hline
4 & D1  & D2  & - & 9.90 & 14.1   \\
5& D1  & D2  & D2 & \textbf{11.1} & \textbf{21.6}   \\
6& D1  & D2  & D1 & 10.5 & 14.4  \\
\end{tabular}
\caption{Results on mixed- and cross-domain experiments.}
\label{tab:cross-domain}
\end{table}

\section{Related works and open issues}
\label{sec:relworks}

The idea of instance-based adaptation exploiting information retrieval dates back to \cite{mahajan1999improved}, in which it was 
developed to dynamically customize a language model for ASR. In statistical MT, it was applied for the same purpose in \cite{eck-etal-2004-language,zhao-etal-2004-language} and later, in  \cite{Hildebrand:2005SHORT}, for domain adaptation.
More recently, different variants of the approach have been proposed for neural MT \cite{farajian2017multi,li-etal-2018-one,zhang-etal-2018-guiding,nonparametric19} and  MT-related tasks  \cite{chatterjee-etal-2017-online}. 
However, differently from ST, all the previously explored translation scenarios involve managing \textit{textual data} for \textit{domain adaptation} purposes. These aspects mark the main differences with our work, which, to the best of our knowledge, is the first attempt to apply instance-based adaptation to cope with data paucity in a speech-related task.

On this front, it is worth remarking that the challenges posed by speech input data can not be addressed with the mere application of  previous text-based techniques. Indeed, differently from MT that only deals with \textit{what} a sentence says in terms of content, the ST (or ASR) input has a more complex nature. 
Together with the conveyed meaning, it also provides  information about the acoustic properties of the spoken utterances (e.g. speaker's voice, recording conditions) describing \textit{how} meaning is expressed. 
This adds additional challenges to instance-based adaptation, where fine-tuning can exploit the retrieval of ``similar'' instances from the  point of view of the audio (e.g. a similar voice), the content (a similar meaning), or both. This paper provides a first exploration along this direction, in which the two aspects are not decoupled. A strand of future works will focus on better understanding and balancing their contribution, as well as dynamically leveraging the notion of similarity (e.g. by a similarity-informed setting of the model's hyper-parameters).

The deeper exploration of different domain-adaptation strategies represents another promising strand of research. In principle, besides maximizing data exploitation in scarce resource conditions, instance-based adaptation 
would allow to simultaneously  manage multiple domains with one single ST system. This is a crucial feature from the industrial standpoint, where training and maintaining domain-dedicated models is costly and time-consuming. We demonstrated the feasibility of the approach with initial experiments but several technical 
aspects still remain to be explored 
(e.g. whether to ``reset'' the model after each update to preserve its performance on all the domains
or to keep the updated one so to favour knowledge transfer across domains when processing new translation requests).

\section{Conclusions}
We proposed a method to maximize data exploitation in the scarce resource conditions posed by end-to-end ST. 
The method is based on 
fine-tuning 
at inference time 
a pre-trained model on a set of instances retrieved from the original training data or from an external corpus based on their similarity with the input audio.
We evaluated our approach in different data conditions (different languages, in/out-of-domain adaptation) reporting coherent improvements over 
generic ST systems and highlighting promising 
research directions for the future.

\section*{Acknowledgements}
This  work  is  part  of  a  project  financially  supported  by  an Amazon AWS ML Grant.

\bibliographystyle{IEEEbib}
\bibliography{strings,refs}

\end{document}